\documentclass{article}
\PassOptionsToPackage{usenames,svgnames,dvipsnames}{xcolor}

 \usepackage[main,nonatbib, final]{neurips_2026}

\usepackage{amsmath,amsfonts,bm}

\def\eqref#1{equation~\ref{#1}}

\def\1{\bm{1}}

\def\vone{{\bm{1}}}

\def\mA{{\bm{A}}}
\def\mB{{\bm{B}}}

\def\mH{{\bm{H}}}
\def\mI{{\bm{I}}}

\def\mK{{\bm{K}}}
\def\mL{{\bm{L}}}

\def\mS{{\bm{S}}}
\def\mT{{\bm{T}}}

\def\mZ{{\bm{Z}}}

\DeclareMathAlphabet{\mathsfit}{\encodingdefault}{\sfdefault}{m}{sl}
\SetMathAlphabet{\mathsfit}{bold}{\encodingdefault}{\sfdefault}{bx}{n}

\usepackage[utf8]{inputenc} 
\usepackage[T1]{fontenc}    
\usepackage{hyperref}       
\usepackage{url}            
\usepackage{booktabs}       
\usepackage{amsfonts}       
\usepackage{nicefrac}       
\usepackage{microtype}      
\usepackage{xcolor}         
\usepackage[numbers]{natbib}
\usepackage{graphicx}
\usepackage{amsmath,amssymb,amsthm,mathtools}

\usepackage{paralist}
\usepackage{enumitem}
\usepackage{bbm}

\definecolor{lightblue}{rgb}{0.88, 0.95, 0.98}
\definecolor{lightpink}{rgb}{0.98, 0.88, 0.95}
\definecolor{superlightred}{rgb}{0.99, 0.92, 0.92}
\definecolor{superlightgreen}{rgb}{0.92, 0.99, 0.92}
\definecolor{darkergreen}{rgb}{0.1, 0.6, 0.1}
\definecolor{squareblue}{HTML}{6ce3cf}
\usepackage{algorithm}
\usepackage{algpseudocode}
\usepackage[table]{xcolor}

\hypersetup{%
  pdfborder = {0 0 0},
  bookmarksdepth = section,
  citecolor = DarkGreen,
  linkcolor = DarkBlue,
  urlcolor = DarkBlue,
  colorlinks = true
}%

\title{Transferring Architecture as an Optimization Prior}

\author{Vighnesh Subramaniam\textsuperscript{1}\thanks{Corresponding author. Code: \url{https://github.com/vsubramaniam851/net-convert}}~\textbf{,}
\textbf{Boris Katz}\textsuperscript{1}\textbf{,} \textbf{Brian Cheung}\textsuperscript{2}\\
$^1$MIT CSAIL, $^2$UCSF
\\
$^{1}$\texttt{\small\{vsub851,boris\}@mit.edu} \\$^2$\texttt{\small bcheung@ucsf.edu}
}

\begin{document}

\maketitle

\begin{abstract}
Initialization determines the optimization ``fate'' of a model, going far beyond just setting the scale of its weights. It establishes a fundamental initial geometry over the input data, dictating certain aspects such as which examples are considered close and which representations gradient descent will naturally refine first. In this paper, we view the initial geometry of a network as an \emph{optimization prior} and ask whether a network can change its optimization prior by referencing a completely different architecture. To investigate this question, we use an initialization procedure where a target network is aligned to the representational geometry of a randomly initialized guide network before any downstream training occurs. Crucially, alignment to the guide network cannot transfer learned features, never receives gradients, and does not use labels. We demonstrate that this cross-architecture transfer of ``fate'' benefits the target network in two key domains: (1) instilling the optimization benefits of depth into shallow networks by aligning them to deeper networks, and (2) transferring representational granularity by using fine-resolution networks to shape the geometry of networks with coarser representations. In both cases, the alignment results in improved downstream performance of the target network. Ultimately, this demonstrates that a model's architectural destiny can be decoupled from its physical structure and explicitly modified at initialization.
\end{abstract}

\section{Introduction}
\begin{figure}
    \centering
    \includegraphics[width=0.95\textwidth]{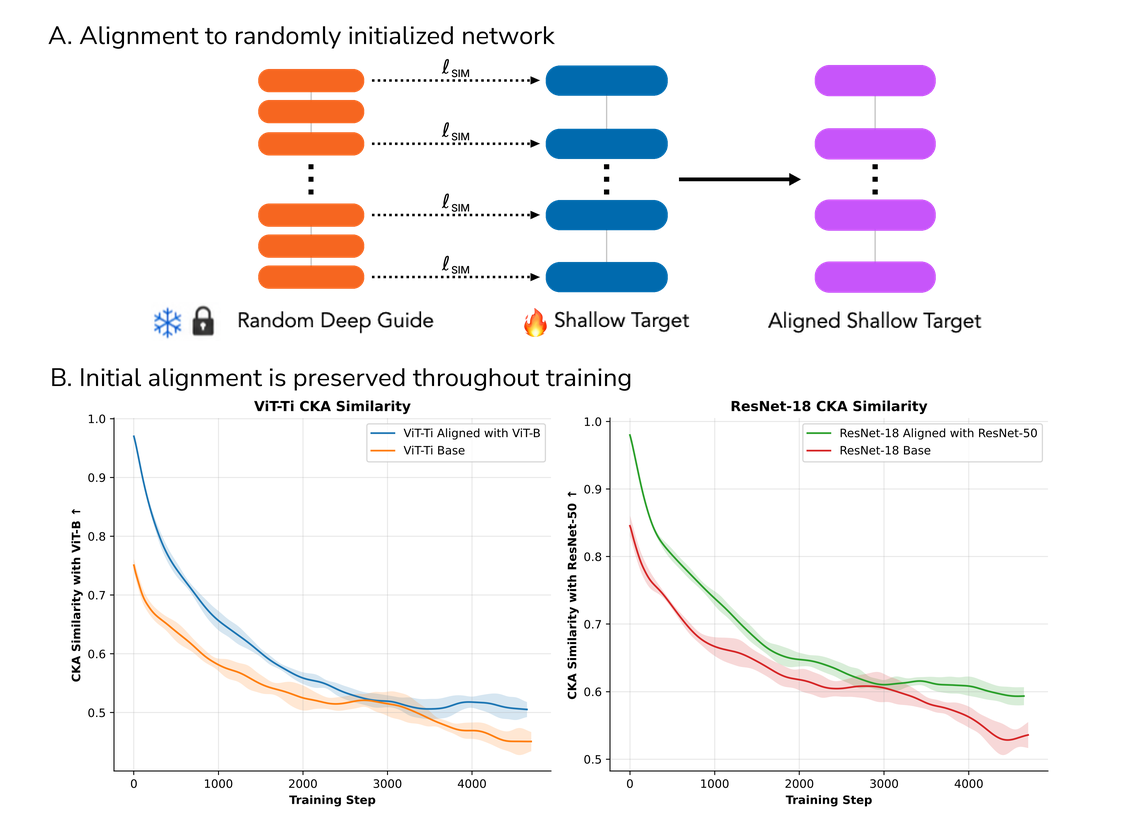}
    \caption{\small \textbf{The Optimization Prior.}
        (A) A frozen, randomly initialized guide defines a representational geometry. We align a target network to this guide at initialization, discard the guide, and then train the aligned target normally.
        (B) During supervised training, aligned ResNet and ViT targets remain closer to their deep guides than standard targets, suggesting that the initialization persists beyond the alignment stage.}
    \label{fig:concept}
    \vspace{-3ex}
\end{figure}

Classical initialization methods traditionally focus on parameter scale to stabilize signal and gradient propagation \citep{glorot2010understanding, he2015delving, arpit2019initialize, mohajerin2017state}. But an initialization also defines something richer, notably, an initial representation of the data. Before a network takes any gradient steps, its random weights already determine geometric relationships among data points such as which examples are close and which relationships govern earliest stages of training. We refer to the geometric bias that an initialization induces on subsequent learning as an \emph{optimization prior}. This perspective raises a question about initialization beyond signal stability, namely what initial geometry a network should induce. 

Architectural choices such as depth and patch resolution change the computations applied to inputs and therefore, the induced geometry. These choices indicate that certain network architectures have better priors than others. We ask whether a network can change its optimization prior by referencing a completely different architecture. This question is motivated by the principle that the initial geometry need not be treated as fixed. In fact, this geometry could be reshaped by referencing the geometry of another architecture via optimization of representation similarity objectives. Representational similarity methods such as SVCCA \citep{raghu2017svcca} and CKA \citep{kornblith2019similarity} were originally developed to compare hidden states across networks, layers, and training runs. More recent work has motivated this perspective by transferring architectural inductive biases across networks \citep{sarnthein2023random, subramaniam2025network, subramaniam2025training}. If we can transfer architecture and reshape geometry, networks with better downstream performance, i.e., ones that use more computations via depth, width, or tokenization, could transfer their optimization priors to networks that are shallower or narrower, leading to downstream performance improvements. More importantly, this has scientific implications for initialization and its connection with architecture. Could architectures with increasingly desirable properties, e.g., deeper, wider, etc., transfer \emph{more} useful geometry to shallow or narrow networks? 

As stated, we formulate the transferring architecture as an optimization prior using representational similarity, designing an initialization-only procedure for choosing the starting geometry of a target network. The procedure is simple. First, we choose a target network, which is the model that will ultimately be trained and deployed. Next, we choose a guide network, which is sampled from a standard random initialization and then frozen. The guide is never trained on the downstream task. Using an unlabeled alignment set, we optimize the target so that its representations become similar to the guide's representations under a representational dissimilarity objective such as DiffKNN \citep{subramaniam2025network} or CKA. Only the target parameters are updated during this stage. Finally, after alignment, we discard the guide and train the aligned target normally on the supervised downstream objective. Thus, the guide does not provide labels, logits, predictions, or a trained function, which distinguishes it from standard distillation. It only provides an initial representational geometry that is absorbed into the target initialization.

We apply the optimization on two settings, \emph{depth transfer} and \emph{representational granularity transfer}. First, for depth transfer, shallow target networks are aligned to deeper randomly initialized guides and then trained on the downstream task. We evaluate this procedure across several architecture families, including ResNets, vision transformers, ConvNeXtV2 models, and GPT-2 language models, showing downstream improvement across all settings. Second, for representational granularity transfer, ViT-B/32, a network with coarse patch tokens, is aligned to ViT-B/16, a network with fine-grained patch tokens, at initialization and then train the ViT-B/32 on the downstream task. Similarly, this leads to downstream improvement. 

We summarize our contributions below:

\begin{compactenum}
    \item We propose transferring architecture as an \emph{optimization prior}, changing the optimization trajectory of a network.
    \item We show that depth can be transferred as an optimization prior by aligning shallow targets to deeper guides across vision and language architectures.
    \item We extend this idea to representational granularity by aligning coarse-grained target representations to fine-grained guide representations, focusing on coarse-token ViT-B/32 targets and fine-token ViT-B/16 guides.
    \item Critically, we find that our results lead to improved downstream performance across all settings. Moreover, even aligning to guide networks with data sampled Gaussian noise leads to improvements in downstream performance.
    \item We analyze the mechanism of transfer by measuring guide proximity during training, demonstrating chained transfer, and deriving how kernel alignment at initialization explains similar early supervised training trajectories.
\end{compactenum}
\vspace{-2ex}
\section{Related Work}
\vspace{-1ex}

We study representational similarity as an initialization-only optimization prior for architectural transfer. This connects to initialization methods that shape trainability before learning and distillation methods that use teachers, including random teachers, to guide students. Our setting differs in that the guide is randomly initialized and discarded after shaping the target initialization.

\textbf{Network Initialization.} Initialization is central to trainability. Classical schemes choose weight distributions to preserve activation and gradient statistics, including Xavier initialization \citep{glorot2010understanding}, He initialization \citep{he2015delving}, and orthogonal initializations motivated by deep linear dynamics \citep{saxe2013exact}. Our work is closest to data-dependent initialization, where weights are chosen using data statistics rather than only layer fan-in, fan-out, or nonlinearities. Examples include data-dependent normalization for CNNs \citep{krahenbuhl2015data}, LSUV initialization \citep{mishkin2015all}, and PCA- or variance-based initialization procedures \citep{koturwar2017weight, shin2020trainability}. A related theoretical literature studies how random initialization shapes signal propagation, expressivity, and trainability in deep networks \citep{poole2016exponential, schoenholz2016deep, pennington2017resurrecting, xiao2018dynamical, zhang2019fixup}. We present a complementary form of data-dependent initialization: the data supply the alignment set, while a random guide architecture supplies the representational geometry.

\textbf{Self-Supervised Learning.} Learning favorable geometry for downstream learning has been a key goal of self-supervised learning (SSL), particularly in computer vision where the goal is to reconstruct representations of image patches. Contrastive methods such as SimCLR align representations of different views of the same image \citep{chen2020simple} while masked autoencoders reconstruct missing image patches \citep{he2022masked}. DINOv2 and DINOv3 combine image-level DINO and patch-level iBOT objectives to build strong self-supervised representations \citep{oquab2023dinov2, simeoni2025dinov3}. DINOv3 further preserves relationships between relationships between through Gram anchoring. In language settings, recent work has explored unsupervised on-policy self-distillation, using a model's own generations to provide its training signal \citep{li2026policy}. In this paper, we apply alignment at initialization, a slightly simpler variant than SSL. Our goal is to show that such alignment reshapes an optimization prior depending on the guide network, implying that such findings could extend to using SSL methods for alignment. 

\textbf{Distillation with Randomly Initialized Networks.} Knowledge distillation trains a student to match signals from a teacher, usually a trained high-capacity model \citep{hinton2015distilling}. Later work distills intermediate representations or attention maps, including FitNets \citep{romero2014fitnets}, attention transfer \citep{zagoruyko2016paying}, and contrastive representation distillation \citep{tian2019contrastive}. Self-distillation and related methods show that the teacher need not be a larger external model; students can learn from identical or evolving teachers, as in Born-Again Networks \citep{furlanello2018born}, BYOL \citep{grill2020bootstrap}, and DINO \citep{caron2021emerging}. Several recent works further show that randomly initialized networks can provide useful learning signals \citep{sarnthein2023random, amid2022learning, dong2023diswot}, and representational alignment methods show that random guides can transfer architectural priors without trained task knowledge \citep{subramaniam2025training, subramaniam2025network}.

Our work builds on this view but differs from distillation in two ways. First, the guide provides no logits, labels, predictions, or trained solution. Second, the guide is absent during downstream supervised training. Thus, the transferred object is an initial representational geometry rather than an ongoing teacher signal.

\section{Representational Alignment Reshapes an Optimization Prior}
\label{sec:theory-main}
Given the goal of transferring architectural properties, we first introduce our procedure for changing an optimization prior using representational alignment objectives and then describe some simple derivations to show that optimizing for alignment changes downstream training. 

\subsection{Optimization Prior Procedure}
Let $S_\theta$ denote the target network whose optimization prior we want to shape, and
let $D_\phi$ denote a guide network. The guide is sampled from its standard random
initialization, frozen, and never trained on labels. Given an unlabeled alignment
set $U$, we define representation readouts
\[
R_S(u;\theta), \qquad R_D(u;\phi),
\]
which extract the target and guide representations to be compared on an input
minibatch $u$. These readouts may correspond to final features, intermediate
features, or a collection of internal activations; the specific choice is an
experimental design decision. The optimization prior is obtained by updating only
the target parameters to minimize a representational dissimilarity,
\[
\mathcal L_{\mathrm{prior}}(\theta)
=
\mathbb E_{u\sim U}
\left[
\Delta_{\mathrm{rep}}
\left(
R_S(u;\theta), R_D(u;\phi)
\right)
\right],
\]
where $\Delta_{\mathrm{rep}}$ can be a global similarity objective such as CKA or
a local-neighborhood objective such as DiffKNN. We denote the resulting aligned
target parameters by $\theta_0^S$. After this alignment stage, the guide
$D_\phi$ is discarded and only the initialized target $S_{\theta_0^S}$ is trained
on the downstream supervised task. Thus the guide does not provide labels,
logits, or a trained solution; it only defines a representational geometry that
is absorbed into the target initialization.

\subsection{Why initialization alignment changes optimization}

Representational alignment changes the representations with which the target begins
supervised training. To see why this matters, consider a supervised batch $(X,Y)$ used after alignment. Let
\[
H_0^S(X)=H_{\theta_0^S}(X), \qquad
H_0^D(X)=H_{\phi}(X)
\]
denote the target and guide feature matrices on this batch, and define their feature
kernels
\[
K_0^S(X)=H_0^S(X)H_0^S(X)^\top,
\qquad
K_0^D(X)=H_0^D(X)H_0^D(X)^\top .
\]
Although alignment is performed on the unlabeled set $U$, downstream optimization
is controlled by the geometry induced on the supervised batch $X$. We therefore measure the relevant mismatch by
\[
\varepsilon_0(X)=\|K_0^S(X)-K_0^D(X)\|_{\mathrm{op}} .
\]
This distinction is important. Representational alignment metrics like CKA may only encourage similar representational geometry, but the first supervised updates depend on the raw feature kernel, or equivalently the initial NTK.

The mechanism is clearest when the classifier head is initialized at zero. Write
the logits as
\[
F=H_\theta(X)V,
\]
with $V_0=0$. Then $F_0=0$, and the first supervised step does not move the backbone because the gradient through the backbone is multiplied by $V_0$. Only the readout moves. If
\[
G_0=\nabla_F\mathcal L(F_0,Y),
\]
then the first logit update is
\[
F_1=-\eta K_0(X)G_0.
\]
Thus the initial feature kernel determines which examples influence one another in
the first gradient step. If the aligned target and guide have similar kernels on $X$, then their first supervised logit updates are close:
\[
\|F_1^S-F_1^D\|_F
\le
\eta\,\varepsilon_0(X)\,\|G_0\|_F .
\]
In words, the optimization prior works by changing the similarity map through
which training examples communicate at the start of learning. The guide is no
longer present during supervised training, but its geometry remains encoded in the
target initialization.

\section{Experiments}
\label{sec:experiments}
We evaluate representational similarity as an initialization-only optimization prior for architectural transfer. In all settings, the guide is random and frozen. The target is aligned on unlabeled data, the guide is discarded, and the target is trained with the same supervised recipe as the standard-initialization baseline. We study two settings: depth transfer from deeper guides to shallower targets, and granularity transfer from fine-resolution ViT guides to coarser ViT targets.

\textbf{Initialization-only alignment via NoT}: In order to test alignment as a prior, we incorporate a mechanism for optimizing alignment. Our alignment procedure follows the stagewise replacement principle of Network of Theseus \citep{subramaniam2025network}. We use NoT only as a practical alignment operator and to validate our scientific findings. Let $r_\phi$ denote a frozen, randomly initialized reference network, and let $s_\theta$ denote the target network whose initialization we optimize. Both networks are decomposed into ordered stages. For an input minibatch $u\in U$, let $R_\ell(u)~ \text{and}~ Z_j(u;\theta)$
denote the activations of reference stage $\ell$ and target stage $j$, respectively. Because the reference and target may have different depths or widths, we use coarse correspondence via a monotone stage map $\pi$ that assigns each aligned target stage $j$ to a reference stage $\pi(j)$. We specify the stage mapping across architectures in Section~\ref{ap:methods}.

For a set of target stages $A$, we optimize
\[
\mathcal L_{\mathrm{align}}(\theta_A)
=
\mathbb E_{u\sim U}
\left[
\frac{1}{|A|}
\sum_{j\in A}
\Delta_{\mathrm{sim}}
\left(
Z_j(u;\theta),\!\left(R_{\pi(j)}(u)\right)
\right)
\right],
\] 

During each alignment phase, only the target parameters in the active stage set $A$ are updated.  After alignment, all target parameters are unfrozen and trained normally on the supervised objective. We use two similarity objectives. We mainly show results with DiffKNN, a differentiable nearest-neighbor similarity objective that compares local neighborhood structure rather than only global kernel alignment. We also show results with linear CKA, which compares global pairwise geometry in a minibatch, but show the full set of results in Section~\ref{ap:cka_performance}. Both objectives operate through batchwise relational structure and therefore allow alignment between stages with different hidden dimensions. More details on both are provided in Section~\ref{ap:methods}.

Unlike the full progressive conversion setting of \citet{subramaniam2025network}, we do not use all stages in every architecture. We find that aligning the upper half of the target network is more effective in our setting. This suggests that, for architectures within the same broad family, much of the transferable representational geometry emerges in later stages. Unless stated otherwise, all experiments align only the upper-half target stages for one epoch per alignment phase. We compare these choices in Section~\ref{ap:alignment_ablation}.

\textbf{Instilling depth}: We first study whether depth can be transferred as an optimization prior. For each architecture family, we choose a randomly initialized guide and align a target from the same family. The aligned target is trained from scratch on the downstream task using the standard supervised objective. Our experiments yields three types of comparisons: deeper-to-shallower transfer, same-architecture alignment, and shallower-to-deeper transfer. The same-architecture setting controls for the effect of alignment alone, while the shallower-to-deeper setting tests whether the direction of transfer matters. For each target architecture, we report the change in downstream performance relative to training the same target from a standard initialization. For vision models, positive values indicate higher ImageNet top-1 accuracy. For language models, negative values indicate lower perplexity. This produces a transfer matrix whose rows are target models and whose columns are guide models.

We evaluate depth transfer across four architecture families. For ResNets, we use ResNet-10, ResNet-18, ResNet-50, ResNet-101, and ResNet-152 \citep{he2016deep}; ResNet-10 follows the standard ResNet design with a convolutional stem, residual stages, progressive downsampling, and global average pooling. For vision transformers, we use ViT-Tiny, ViT-Small, ViT-Base, and ViT-Large \citep{dosovitskiy2020image}, all with $16\times16$ patches unless otherwise stated. ViT-Tiny has six transformer encoder blocks with hidden size 384, and ViT-Small has eight blocks with hidden size 512. For ConvNeXtV2, we use Atto, Nano, Tiny, Base, and Large variants following the original depth and width configurations \citep{woo2023convnext}. For language modeling, we use GPT-2-Nano, GPT-2-Tiny, GPT-2-Base, GPT-2-Medium, and GPT-2-Large \citep{radford2019language}. GPT-2-Nano and GPT-2-Tiny are defined for this work with four and eight transformer decoder blocks, respectively, and all variants share the same tokenizer, positional encoding scheme, and transformer block structure.

\textbf{Instilling detail}: We next ask whether the same initialization prior can transfer representational granularity. In vision transformers, patch size controls token granularity. A ViT-B/16 represents an image with a finer spatial grid than a ViT-B/32, while ViT-B/32 uses larger patches and therefore forms coarser token representations. We use this contrast to test whether fine-token geometry can shape the initialization of a coarse-token model.

The guide is a randomly initialized ViT-B/16 and the target is a ViT-B/32. We align the target to the guide, then train the ViT-B/32 target normally. Improvement over a standard training baseline indicates that the fine-token guide provides useful initial geometry even though the final deployed target remains a coarse-token model. We compare ViT-B/16 and ViT-B/32 by flattening token representations and aligning class-token representations, allowing comparison of their random global geometry. As with the depth-transfer setting, we add controls for same-tokenization and for the reverse direction, from coarse tokenization to fine tokenization.

\textbf{Datasets, training and evaluation protocol}: For vision experiments, we train on ImageNet-1K and report top-1 accuracy on the validation set \citep{deng2009imagenet}. For language experiments, we train on WikiText-103 and report test set perplexity \citep{merity2016pointer}. We use the standard training, validation, and test splits for each dataset. 

All alignment phases are optimized with AdamW \citep{loshchilov2017decoupled}. Each alignment phase runs for one epoch over the alignment set. After alignment, the target is trained with standard supervised training for 100 epochs with AdamW. The baseline for each target architecture is the same model trained from a standard initialization for 101 epochs with the same optimizer, scheduler, and learning rate used after alignment. Thus, aligned and baseline targets receive the same training budget, optimizer, scheduler, augmentations, and hyperparameters. For all supervised training runs, we use cosine learning-rate decay with 5\% warmup \citep{loshchilov2016sgdr}. Hyperparameters are discussed in Section~\ref{ap:methods}. For ImageNet models, we use the same preprocessing and augmentation policy for aligned and baseline targets. For WikiText-103, we use the same tokenizer, context length, batching procedure, and evaluation protocol for aligned and baseline targets.

\section{Results}
We separate our results into findings over transferring depth, transferring detail, and general analyses. For each case, aligned performance is compared to the corresponding baseline target trained from a standard initialization. All reported deltas are averaged over five random seeds, and raw aligned and baseline performances with standard deviations are reported in Appendix~\ref{ap:network_performance}.
\begin{figure}[t]
    \centering
    \includegraphics[width=0.94\textwidth]{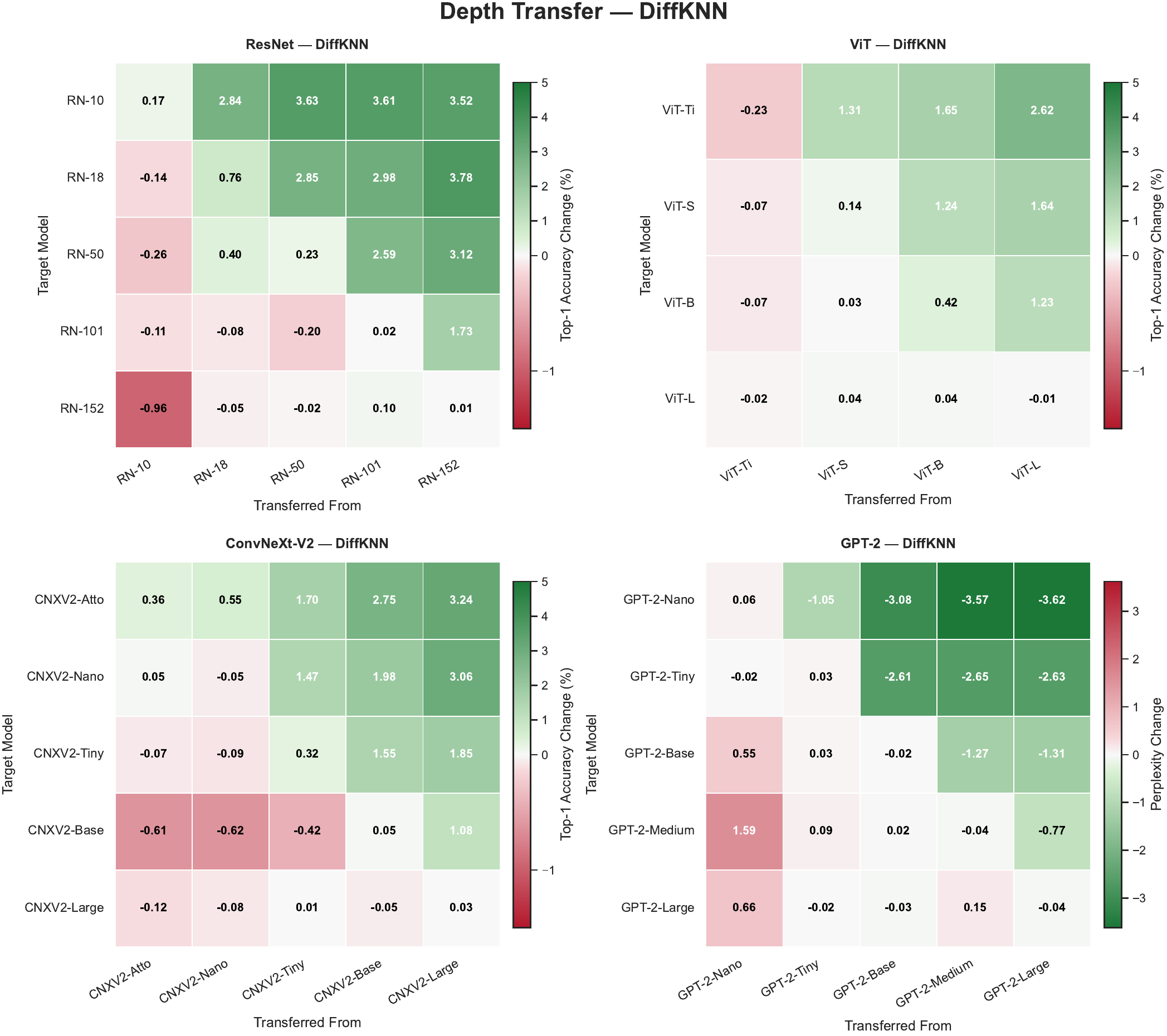}
    \caption{\small \textbf{Depth transfers to shallow networks.}
    We align different networks across different depths at initialization. Rows are targets and columns are random guides used only during alignment. Each cell reports the either the average ImageNet top-1 accuracy or Wikitext-103 test set perplexity change relative to the same target trained from standard initialization. Green is better (higher for ImageNet and lower for GPT-2). Across ResNets, ViTs, ConvNeXtV2, and GPT-2 models, deeper guides consistently improve shallower targets. Raw performance with error bars is reported in Section~\ref{ap:network_performance} and performance with CKA is shown in Section~\ref{ap:cka_performance}.}
    \label{fig:depth_transfer}
    \vspace{-1ex}
\end{figure}

\begin{figure}
    \centering
    \includegraphics[width=0.9\textwidth]{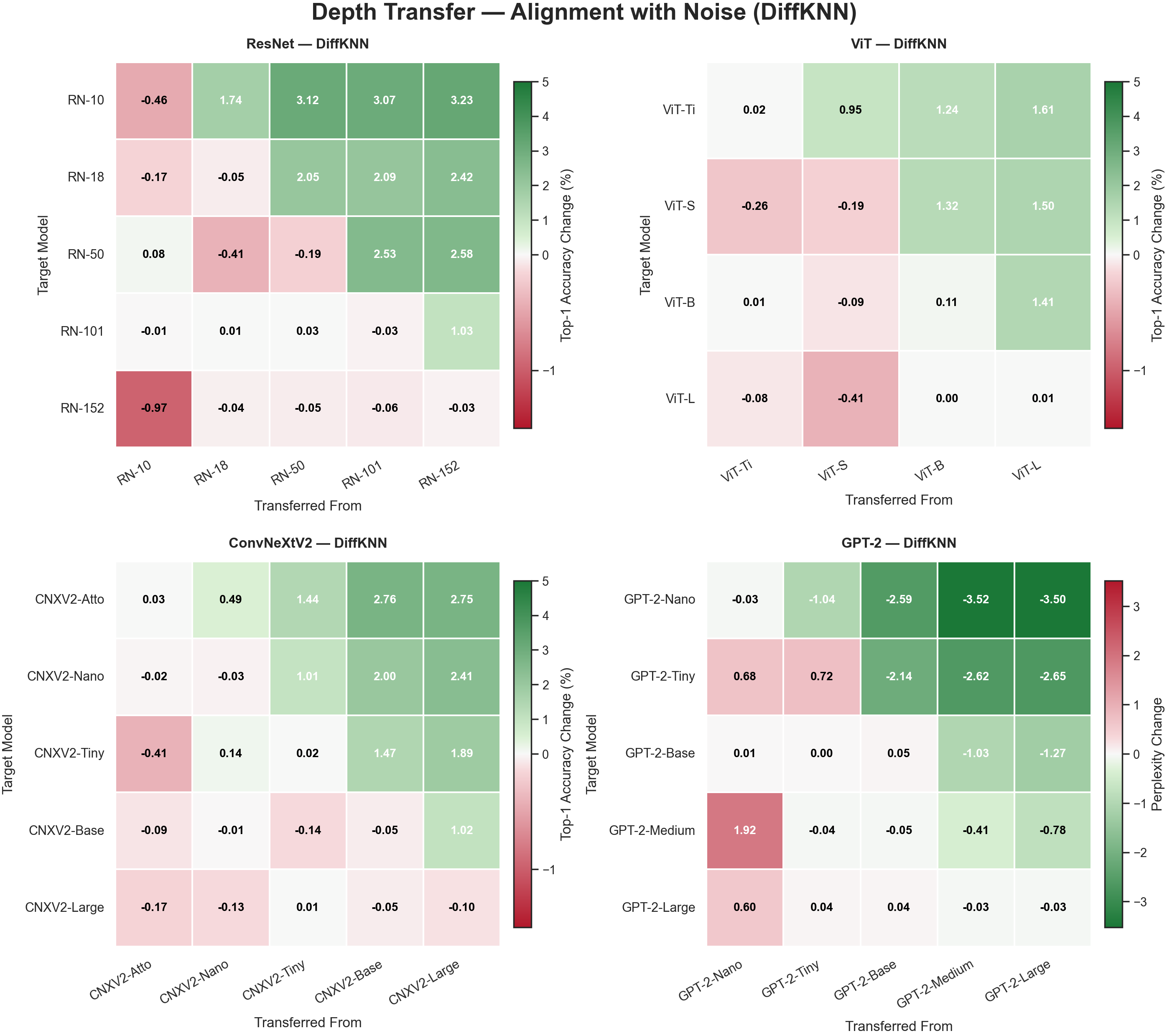}
    \caption{\small \textbf{Transferring architecture without data}. We align target networks to guide networks without relying on real data. Data is sampled from a Gaussian distribution for both images and text tokens. Gains over baseline training are observed for all target networks. This demonstrates that target networks can have their optimization priors reshaped just by fitting the geometric distribution of the guide. Gains were not driven by data statistics.}
    \label{fig:noise}
\end{figure}


\textbf{Instilling Depth}: We first measure our ability to instill the geometry of deeper architectures into shallower architectures by optimizing representational alignment at initialization. We show results for vision networks in Figure~\ref{fig:depth_transfer}. Across both settings and with both similarity metrics, we observe that deeper guide networks lead to better target performance. In vision models, deeper ResNets, ViTs, and ConvNeXtV2 networks improve ImageNet accuracy after alignment at initialization. In language models, deeper GPT-2 guides improve shallow GPT-2 perplexity on WikiText-103. The strongest improvements reach roughly four top-1 accuracy points for vision or four perplexity points for language. Generally, DifKNN and CKA give similar improvements for depth transfer. When the guide is shallower than the target, performance usually stays close to the baseline or degrades slightly. These reverse-direction controls show that alignment is a partial constraint on target geometry, not an architectural projection.  A shallow guide seems to perturb a deeper target network slightly, but it does not force the target to collapse to shallow-model behavior, because the guide is removed before supervised training and the target retains its own parameterization. When the guide and target have the same architecture, alignment also has little effect, with only small improvements or degradations in accuracy or perplexity.

The strongest pattern is that target performance improves as guide depth increases. For example, aligning ResNet-10 to a randomly initialized ResNet-152 gives stronger downstream performance than aligning ResNet-10 to ResNet-50 or ResNet-101. Similarly, the shallow GPT-2 variants benefit most from the largest GPT-2 guides, with GPT-2-Large and GPT-2-Medium producing the strongest perplexity improvements. This suggests a possible initialization scaling law, where guide networks that apply more computation induce more useful initial geometry for the target.

\textbf{Alignment without data}: Currently, our alignment procedure relies on real data for optimization. This begs the question of whether real, natural images or text are necessary to see the gains. If we can establish that target networks can improve by aligning to a deeper network without using real data, this motivates our approach as a method closer to weight initialization. To test this, target networks are optimized to align to guide networks using synthetic data. To generate synthetic data, we can sample data from a Gaussian. Either images or token IDs are sampled from Gaussian distributions as input data to the image or language models we align. Results are shown in Figure~\ref{fig:noise}. We show that target networks still improve even when alignment does not use real data. This is significant, demonstrating that gains are not driven by data statistics.

\begin{figure}
    \centering
    \includegraphics[width=0.93\textwidth]{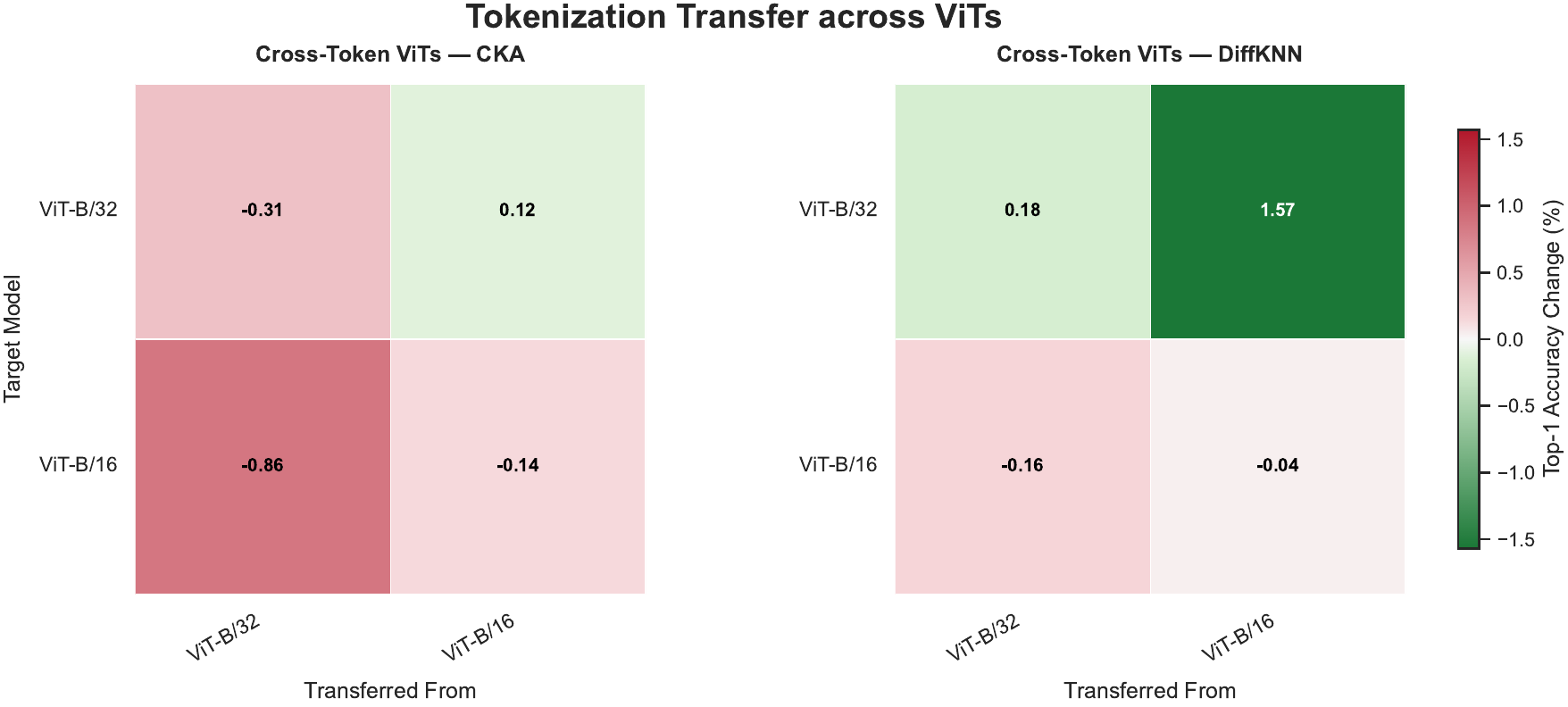}
    \caption{\small \textbf{Detail can transfer to coarse vision transformers.} 
    Rows are target tokenizations, and columns are randomly initialized guide tokenizations. Each cell reports ImageNet top-1 accuracy change relative to the same target trained from standard initialization; green indicates improvement. The top-right cell tests fine-to-coarse transfer by aligning a coarse-token ViT-B/32 target to a fine-token ViT-B/16 guide. DiffKNN gives the clearest improvement for transferring fine-token geometry into ViT-B/32, while CKA provides a weaker signal when the random ViT-B/16 and ViT-B/32 representations are already highly aligned.}
    \label{fig:tokenization_transfer}
\end{figure}

\textbf{Instilling detail.}
We next test whether fine-resolution ViT geometry can initialize a coarser ViT. Figure~\ref{fig:tokenization_transfer} aligns ViT-B/32 targets to ViT-B/16 guides, with same-tokenization and reverse-direction controls. CKA gives little improvement, while DiffKNN substantially improves the ViT-B/32 target. The same-tokenization and reverse-direction controls are much weaker, suggesting that fine-token geometry provides a useful initialization signal for coarse-token representations.

This setting also clarifies the difference between CKA and DiffKNN. The initial CKA between random ViT-B/32 and ViT-B/16 is already about $0.95$, leaving little room for optimization. DiffKNN is less saturated because it compares local neighborhoods, allowing it to distinguish tokenizations that have similar global kernel structure. Thus local geometry can be more informative than global kernel shape when transferring representational granularity.

\begin{figure}
    \centering
    \includegraphics[width=0.93\textwidth]{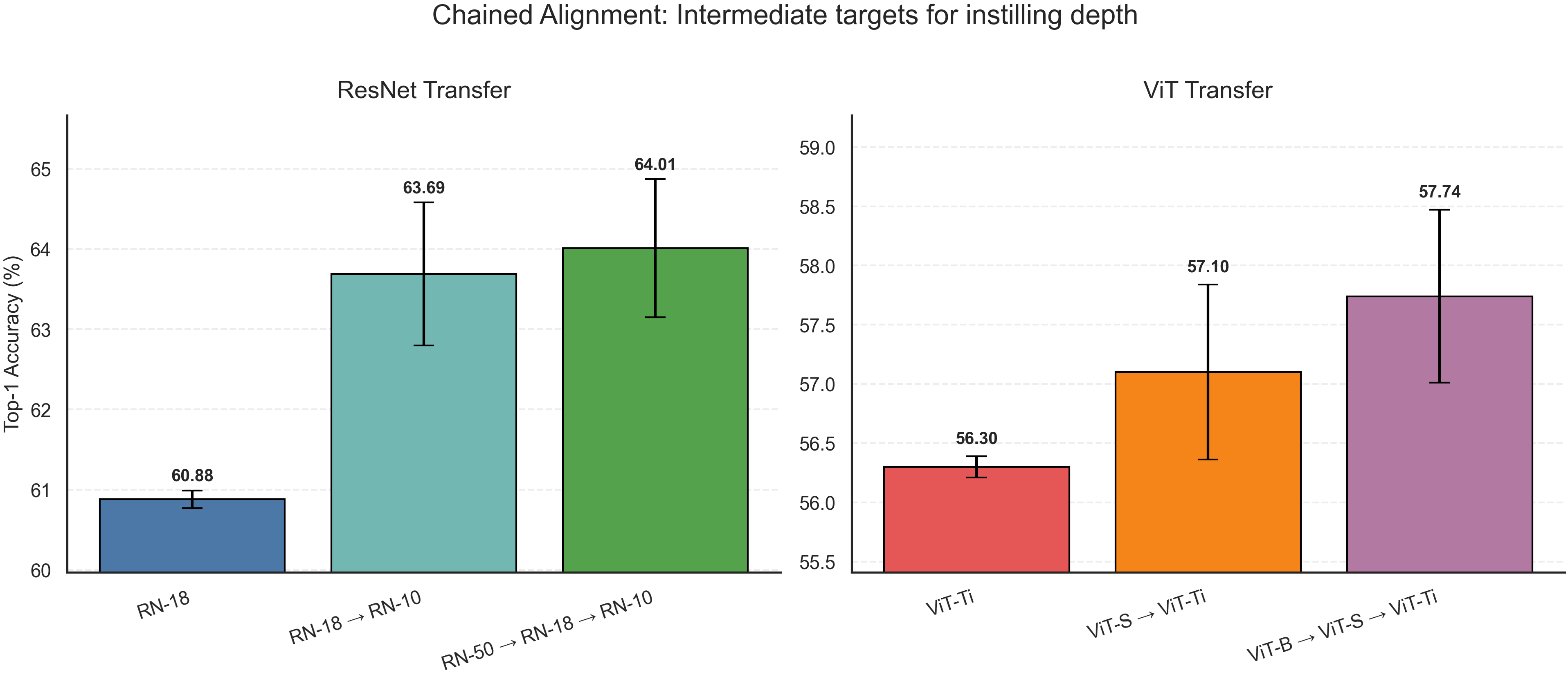}
    \caption{\small \textbf{Chained transfer improves shallow targets through intermediate networks.} We first align an intermediate target to a deep guide, then use that aligned intermediate model as the guide for a smaller target. For example, ViT-S is aligned to ViT-B, and the resulting ViT-S is then used to align ViT-Ti. Chained transfer improves over using a standard intermediate guide, suggesting that guide-induced geometry can be passed through successive target networks.}
    \label{fig:chained_transfer}
\end{figure}

\textbf{Chained transfer.}
We also ask whether transferred geometry can pass through intermediate targets. Figure~\ref{fig:chained_transfer} first aligns an intermediate model to a deep guide, then uses the aligned intermediate model as the guide for a smaller target. For example, ViT-S is aligned to ViT-B and then used to align ViT-Ti. This suggests that guide-induced geometry is partially compositional: it can be absorbed by an intermediate target and then passed to a smaller model.

\begin{figure}
    \centering
    \includegraphics[width=0.93\textwidth]{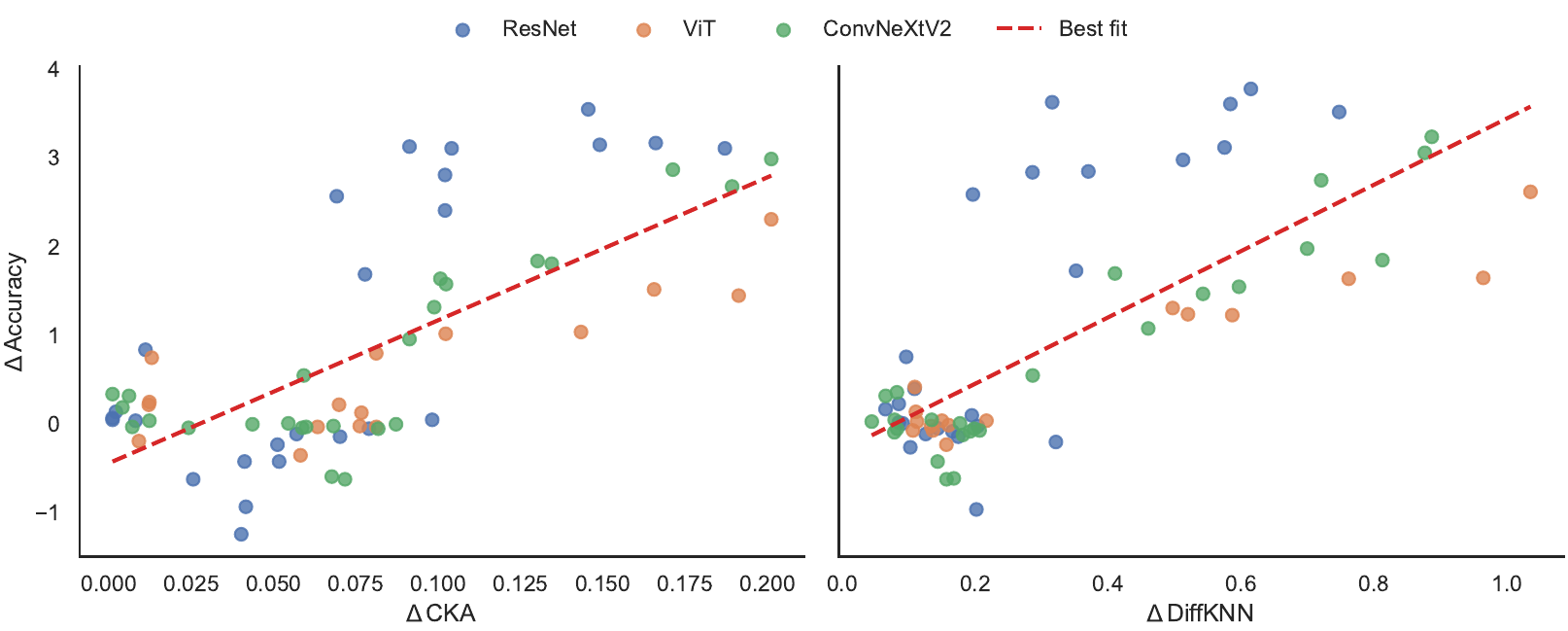}
    \caption{\small \textbf{Alignment movement predicts downstream gains.}
    For each guide-target pair, we compare the change in alignment during the initialization stage, $\Delta \mathrm{sim}=\mathrm{sim}_{\mathrm{end}}-\mathrm{sim}_{\mathrm{start}}$, with the final ImageNet top-1 improvement of the aligned target. Larger alignment movement generally corresponds to larger downstream gains. When CKA is already saturated, the target moves little and gains are smaller; DiffKNN remains informative by measuring local neighborhood structure.}
    \label{fig:sim_acc}
\end{figure}

\textbf{Alignment movement vs. final accuracy}: As suggested by the detail-transfer results, one factor that affects transfer is how much the target geometry actually moves during initialization alignment. For each guide-target pair, we measure the starting alignment $a_{\mathrm{start}}$, the ending alignment $a_{\mathrm{end}}$, and the alignment movement $\Delta a=a_{\mathrm{end}}-a_{\mathrm{start}}$. We compare $\Delta a$ to downstream improvement in Figure~\ref{fig:sim_acc}. Larger alignment movement generally corresponds to larger accuracy gains, suggesting that the optimization prior is strongest when the alignment objective induces a nontrivial change in target geometry. When CKA is already close to saturated, as in the ViT-B/16 to ViT-B/32 setting, alignment moves little and provides a weaker signal. DiffKNN is less saturated because it compares local neighborhood structure, allowing it to move the target in settings where global kernel geometry is already similar. Together, these results suggest that CKA and DiffKNN are complementary, and representational alignment is a partial constraint on geometry rather than an architectural projection.
\section{Conclusion}
We show that the geometry of a network at initialization, i.e. its \emph{optimization prior}, can be changed through representational alignment with another, potentially more desirable architecture. Across vision and language settings, randomly initialized guides transfer useful architectural structure without labels or trained task knowledge. In particular, depth can be instilled into shallow targets, and fine-resolution representational geometry can be transferred into coarser ViT representations. Aligned targets remain closer to their guides during training, transfer can be chained through intermediate networks, and the induced initial kernel explains the earliest supervised dynamics.

Our results suggest a broader view of initialization and architecture. Rather than treating architecture only as a fixed computation graph, we show that some of its geometric consequences can be used transiently to shape how another network begins learning. This gives random architectures a scientific role as probes for optimization biases such as depth, locality, recurrence, and representational granularity. It also suggests new initialization strategies based on composing guide networks, chaining intermediate targets, and combining global and local similarity objectives such as CKA and DiffKNN. More broadly, representational alignment offers a way to separate the optimization benefits of architectural structure from the requirement that this structure persist in the final model.

\textbf{Limitations.}
Our study does not exhaust the space of possible architectures, datasets, or alignment objectives.
The improvements are consistent but modest, with the stron9gest gains around three to four accuracy points for vision or perplexity points for language in our tested settings.
The tokenization experiments also show that transfer can be sensitive to the starting similarity between guide and target, especially when CKA is already saturated.
Future work should study stronger local alignment objectives, combinations of CKA and DiffKNN, and additional architectural properties such as width, scale, recurrence, and locality.

\section*{Acknowledgements}
This work was supported by the DARPA Mathematics for the DIscovery of ALgorithms and Architectures (DIAL) program, the NSF award 2124052, the MIT-IBM Watson AI Lab, the Department of the Air Force Artificial Intelligence Accelerator under Cooperative Agreement Number FA8750-19-2-1000, and the Air Force Office of Scientific Research (AFOSR) under award number FA9550-21-1-0399. B.C. is supported by the DARPA Network of Optimal Dynamic Energy Signatures (NODES) program under Agreement No. HR0011269E076. The views and conclusions contained in this document are those of the authors and should not be interpreted as representing the official policies, either expressed or implied, of the Department of the Air Force or the U.S. Government. The U.S. Government is authorized to reproduce and distribute reprints for Government purposes notwithstanding any copyright notation herein.

We would like to thank all reviewers and the area chair of this NeurIPS submission for their thoughtful feedback. We would also like to thank Chris Wang, Kushagra Tiwary, David Mayo, Abdulrahman Alabdulkareem, the Isola Lab at MIT CSAIL, and the Efros Lab at UC Berkeley for their extremely useful input on this work. 

\bibliographystyle{plainnat}
\bibliography{neurips_2026}

\newpage 
\appendix
\section{Appendix Overview}
We cover methodological and experimental details, analyses, results, and ablations. 
In Section~\ref{ap:methods}, we cover additional details for applying NoT \citep{subramaniam2025network} in our work. We describe the alignment metrics used in more detail, our approach for mapping layers across networks, and training details. In Section~\ref{ap:cka_performance}, we show results for our depth transfer setting with CKA. In Section~\ref{ap:network_performance}, we provide raw performance values across our settings to better ground our findings. In Section~\ref{ap:theory}, we expand on our theoretical derivations introduced in the paper. We explore a new transfer setting, recursive transfer, in Section~\ref{ap:recursive}, where we can simulate depth using a single network module applied recursively. In Section~\ref{ap:alignment_ablation}, we provide ablation experiments to justify experimental design choices made in the paper. 

\section{Methodological and Experimental Details}
\label{ap:methods}
\subsection{Similarity Metrics}
\textbf{CKA}: We use linear CKA as our global representational similarity objective.
CKA uses kernel functions on centered representations to compute representational similarity matrices, which are then compared through the Hilbert--Schmidt Independence Criterion (HSIC).
More specifically, suppose we have two sets of representations $\mA\in\mathbb{R}^{b\times d_1}$ and $\mB\in\mathbb{R}^{b\times d_2}$, where $b$ is the minibatch size.
Before computing CKA, we $\ell_2$-normalize each row of $\mA$ and $\mB$.
We then compute the linear Gram matrices
\begin{equation}
    \label{eq:gram}
    \mK = \mA\mA^\top, \qquad \mL = \mB\mB^\top .
\end{equation}

We center the Gram matrices using the centering matrix $\mH$, where
\[
    \mH = \mI_b - \frac{1}{b}\vone\vone^\top .
\]
This gives
\begin{equation}
    \tilde{\mK} = \mH\mK\mH, \qquad
    \tilde{\mL} = \mH\mL\mH .
\end{equation}

We compute HSIC as the Frobenius inner product between the centered Gram matrices,
\begin{equation}
    \operatorname{HSIC}(\mK,\mL)
    =
    \langle \tilde{\mK}, \tilde{\mL}\rangle_F
    =
    \operatorname{tr}(\tilde{\mK}\tilde{\mL}) .
\end{equation}

Finally, we define linear CKA as
\begin{equation}
    g_{\mathrm{CKA}}(\mA,\mB)
    :=
    \operatorname{CKA}(\mK,\mL)
    =
    \frac{
        \operatorname{HSIC}(\mK,\mL)
    }{
        \sqrt{
            \operatorname{HSIC}(\mK,\mK)
            \operatorname{HSIC}(\mL,\mL)
        }
    } .
\end{equation}

In our setting, we optimize representational \emph{dissimilarity}.
Since linear CKA is a similarity score, with larger values indicating more similar representations, we minimize the negative-log CKA:
\begin{equation}
    \Delta_{\mathrm{CKA}}(\mA,\mB)
    =
    -\log(g_{\mathrm{CKA}}(\mA,\mB)).
\end{equation}
Thus, minimizing $\Delta_{\mathrm{CKA}}$ encourages the target and guide representations to have similar centered kernel geometry.

\textbf{DiffKNN}:
We also use a differentiable nearest-neighbor objective, DiffKNN, to compare local neighborhood structure.
The goal of DiffKNN is to make the target network assign high similarity to the same local neighbors as the guide network.

Suppose we have target representations $\mA\in\mathbb{R}^{b\times d_1}$ and guide representations $\mB\in\mathbb{R}^{b\times d_2}$.
We first compute pairwise similarity matrices over examples in the minibatch.
For cosine similarity, we normalize rows and compute
\begin{equation}
    \mS^{A}_{ij}
    =
    \left\langle
        \frac{\mA_i}{\|\mA_i\|_2},
        \frac{\mA_j}{\|\mA_j\|_2}
    \right\rangle,
    ~~
    \mS^{B}_{ij}
    =
    \left\langle
        \frac{\mB_i}{\|\mB_i\|_2},
        \frac{\mB_j}{\|\mB_j\|_2}
    \right\rangle .
\end{equation}
We mask the diagonal entries so that an example cannot select itself as a nearest neighbor.

For each anchor example $i$, we compute the guide's top-$k$ nearest-neighbor set
\begin{equation}
    \mathcal{N}_k^B(i)
    =
    \operatorname{TopK}_{j\neq i}\, \mS^B_{ij}.
\end{equation}
This set is computed from the frozen guide network and is treated as a fixed target during the update.
We then construct a binary neighborhood target matrix $\mT\in\{0,1\}^{b\times b}$:
\begin{equation}
    \mT_{ij}
    =
    \mathbbm{1}\{j\in \mathcal{N}_k^B(i)\},
    \qquad
    \mT_{ii}=0 .
\end{equation}

The target similarities $\mS^A$ are used as logits.
Before applying the loss, we row-standardize the non-diagonal logits.
Let
\begin{equation}
    \mu_i
    =
    \frac{1}{b-1}\sum_{j\neq i}\mS^A_{ij},
    ~~
    \sigma_i^2
    =
    \frac{1}{b-1}\sum_{j\neq i}(\mS^A_{ij}-\mu_i)^2 .
\end{equation}
For $j\neq i$, we define
\begin{equation}
    \mZ^A_{ij}
    =
    \frac{\mS^A_{ij}-\mu_i}{\tau(\sigma_i+\epsilon)},
\end{equation}
where $\tau$ is a temperature parameter.
The diagonal entries are masked out and do not contribute to the loss.

We optimize a binary cross-entropy objective that encourages the target to assign high probability to the guide's nearest neighbors and low probability to other examples.
For each pair $(i,j)$, define
\begin{equation}
    \ell_{ij}
    =
    -
    \mT_{ij}\log \sigma(\mZ^A_{ij})
    -
    (1-\mT_{ij})\log\left(1-\sigma(\mZ^A_{ij})\right),
\end{equation}
where $\sigma(\cdot)$ is the logistic sigmoid.
For each anchor $i$, we average the positive and negative terms separately:
\begin{equation}
    \ell_i^+
    =
    \frac{1}{k}
    \sum_{j\neq i}
    \mT_{ij}\ell_{ij},
    ~~
    \ell_i^-
    =
    \frac{1}{b-1-k}
    \sum_{j\neq i}
    (1-\mT_{ij})\ell_{ij}.
\end{equation}
The DiffKNN dissimilarity is then
\begin{equation}
    \Delta_{\mathrm{DiffKNN}}(\mA,\mB)
    =
    \frac{1}{b}
    \sum_{i=1}^{b}
    \frac{
        \ell_i^+
        +
        \lambda \ell_i^-
    }{
        1+\lambda
    },
\end{equation}
where $\lambda$ controls the weight of negative examples.
In our experiments, the guide network is frozen, so gradients are taken only through the target logits $\mZ^A$.
Thus, DiffKNN provides a differentiable objective for matching the guide's local neighborhood structure, even though the guide neighborhoods themselves are computed using a discrete top-$k$ operation.

Unless otherwise stated, we use cosine similarity with row-standardized logits, $\lambda=1$, and fixed values of $k$ and $\tau$ across each experiment.
In all experiments, we set $k$ to 10 and $\tau$ to 1.0.

\subsection{Stage Mapping}
We describe how stages and layers are mapped across target and guide networks.
As covered in Section~\ref{sec:experiments}, we only used the top half of all networks as justified later in Section~\ref{ap:alignment_ablation}. 
In each setting, we first take a linear mapping from the target network to the layers of the guide network.
This follows a simple linear interpolation where the target network layers are spread out over the guide network layers, similar to the mapping used in \citet{subramaniam2025training}.
We then drop the bottom half of target network layers to get the final NoT stages.

For certain settings, we align different layers.
In ResNets, batch normalization can inflate alignment. 
Therefore, we only consider alignment from the last convolutional layer in each block.
Similarly, we take the pre-layer-normalization layers in ViTs and GPT-2.
This corresponds with the second feedforward layer. 
In ConvNeXt-V2, this corresponds to the last 1x1 convolutional in a ConvNeXt-V2 block.
Each layer represents a stage and is mapped to the corresponding guide network layer.

\subsection{Training Details}
\label{ap:training}
We cover training hyperparameters in Table~\ref{tab:hyperparams}. 
As stated in Section~\ref{sec:experiments}, our optimization via NoT uses one epoch per progressive alignment stage. 
In our setting, we have anywhere between 3-6 alignment stages, making our alignment much shorter than used in previous work. 

Our alignment uses the AdamW optimizer with a tuned learning rate. 
In general, we find one epoch sufficient to get the best alignment optimization. We also use a cosine scheduler \citep{loshchilov2016sgdr} with 3\% warmup. 
When applying standard training, we find that the same learning rate is sufficient to find similar results as what is reported in previous works. 
We apply 100 epochs of standard training with the AdamW optimizer, a cosine scheduler, and 5\% warmup.
Across all training we use gradient clipping to a norm of 1.0 to stabilize training.
Similarly, across all settings, we use a batch size of 256.
For this work, we trained all networks on eight H200s over the period of one month. 
A single conversion can be completed with one H200 in a single day.

\begin{table}
\resizebox{\textwidth}{!}{%
\begin{tabular}{lcccccccc}
\toprule
Setting    & Alignment Epochs per Stage & Optimizer & Learning Rate (Alignment) & Learning Rate (Standard Training) & Warmup (Alignment) & Warmup (Standard Training) & Weight Decay & Gradient Clipping \\ \midrule
ResNets    & 1                          & AdamW \citep{loshchilov2017decoupled}     & $1\times10^{-3}$          & $1\times10^{-3}$                  & 3\%                & 5\%                        & 0.1          & 1.0               \\
ViTs       & 1                          & AdamW     & $1\times10^{-3}$          & $1\times10^{-3}$                  & 3\%                & 5\%                        & 0.1          & 1.0               \\
ConvNeXtV2 & 1                          & AdamW     & $1\times10^{-3}$          & $1\times10^{-3}$                  & 3\%                & 5\%                        & 0.1          & 1.0               \\
GPT-2      & 1                          & AdamW     & $5\times10^{-4}$          & $5\times10^{-4}$                  & 3\%                & 5\%                        & 0.1          & 1.0               \\ \bottomrule
\end{tabular}%
}
\caption{\small\textbf{Training details for depth and detail transfer.} We report all hyperparameters for our optimizer and scheduler. All alignment stages use 1 full epoch of training with 3\% warmup per stage. We use the AdamW optimizer with the same learning rate for alignment and standard training. The learning rate is chosen for best reproduction of baseline performance.
}
\label{tab:hyperparams}
\end{table}

\section{CKA Improvement}
\begin{figure}
    \centering
    \includegraphics[width=0.8\textwidth]{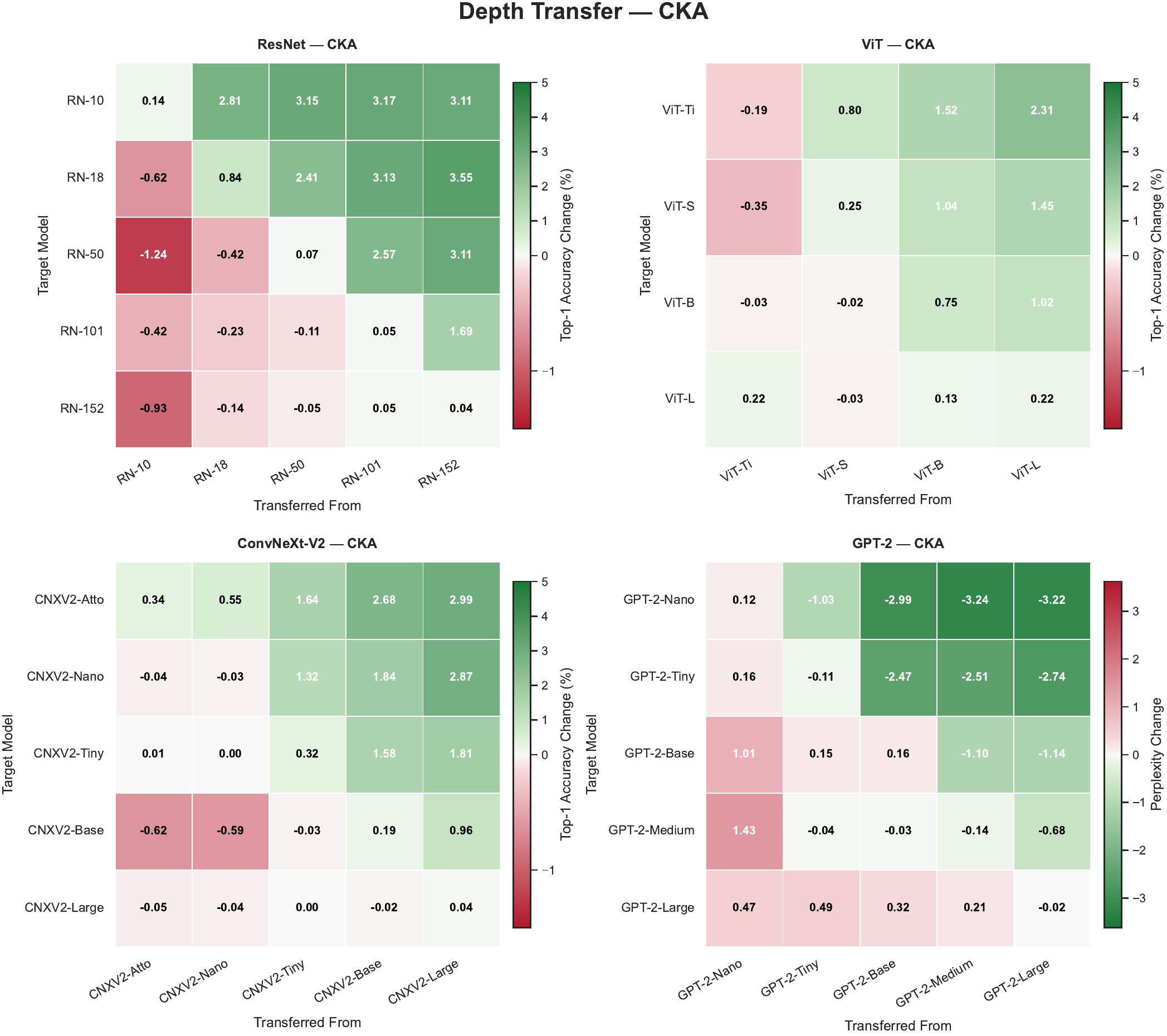}
    \caption{\textbf{Depth transfers with CKA}: We recreate Figure~\ref{fig:depth_transfer} with CKA and find similar trends. Deeper networks transfer more useful initial geometry to shallow networks. This useful geometry translates to use gains on downstream tasks. In general, we find DiffKNN leads to slightly stronger performance}
    \label{fig:cka_depth}
\end{figure}

We show CKA improves on depth transfer in Figure~\ref{fig:cka_depth}. This reproduces Figure~\ref{fig:depth_transfer} and the trends discussed previously.

\label{ap:cka_performance}
\label{ap:network_performance}
\section{Network Performance}

We show raw network performance across our settings in Table~\ref{tab:scale_transfer_compact}. 
This includes the raw performance, accuracy or perplexity, along with the standard error for measuring performance. 
\begin{table*}[t]
\centering
\scriptsize
\setlength{\tabcolsep}{3.2pt}
\renewcommand{\arraystretch}{1.08}

\resizebox{\textwidth}{!}{%
\begin{tabular}{lccccccccccc}
\toprule
\multicolumn{12}{c}{\textbf{ImageNet Top-1 Accuracy} ($\uparrow$): \textbf{ResNet}} \\
\midrule
Target & Baseline & \multicolumn{5}{c}{CKA guide $\rightarrow$ target} & \multicolumn{5}{c}{DiffKNN guide $\rightarrow$ target} \\
\cmidrule(lr){3-7}\cmidrule(lr){8-12}
 &  & RN-10 & RN-18 & RN-50 & RN-101 & RN-152 & RN-10 & RN-18 & RN-50 & RN-101 & RN-152 \\
\midrule
RN-10  & 60.88 $\pm$0.59 & 61.02$\pm$0.92 & 63.69 $\pm$0.96 & 64.03 $\pm$0.99 & 64.05 $\pm$0.56 & 63.99 $\pm$0.49 & 61.05$\pm$0.54 & 63.72$\pm$0.61 & 64.51$\pm$0.51 & 64.49$\pm$0.89 & 64.40$\pm$0.92 \\
RN-18  & 68.16$\pm$0.11 & 67.54 $\pm$0.81 & 69.00 $\pm$0.44 & 70.57 $\pm$0.89 & 71.29$\pm$0.65 & 71.71$\pm$0.74 & 68.02$\pm$0.47 & 68.92$\pm$0.93 & 71.01$\pm$0.26 & 71.14$\pm$0.46 & 71.94$\pm$0.54 \\
RN-50  & 74.43$\pm$0.14 & 73.19$\pm$0.75 & 74.01$\pm$0.72 & 74.50$\pm$0.61 & 77.00$\pm$0.696 & 77.54$\pm$0.78 & 74.17$\pm$0.86 & 74.83$\pm$0.81 & 74.66$\pm$0.62 & 77.02$\pm$0.71 & 77.55$\pm$0.86 \\
RN-101 & 76.11$\pm$0.16 & 75.69$\pm$0.78 & 75.88$\pm$0.71 & 76.00$\pm$0.51 & 76.16$\pm$0.72 & 78.80$\pm$0.69 & 76.00$\pm$0.61 & 76.03$\pm$0.54 & 75.91$\pm$0.65 & 76.13$\pm$0.52 & 78.84$\pm$0.78 \\
RN-152 & 77.05$\pm$0.12 & 76.12$\pm$0.91 & 76.91$\pm$0.96 & 77.00$\pm$1.01 & 77.10$\pm$0.85 & 77.09$\pm$1.02 & 76.09$\pm$0.56 & 77.00$\pm$0.67 & 77.03$\pm$1.00 & 77.15$\pm$0.99 & 77.06$\pm$0.87 \\
\bottomrule
\end{tabular}%
}

\vspace{0.75em}

\resizebox{\textwidth}{!}{%
\begin{tabular}{lccccccccc}
\toprule
\multicolumn{10}{c}{\textbf{ImageNet Top-1 Accuracy} ($\uparrow$): \textbf{ViT}} \\
\midrule
Target & Baseline & \multicolumn{4}{c}{CKA guide $\rightarrow$ target} & \multicolumn{4}{c}{DiffKNN guide $\rightarrow$ target} \\
\cmidrule(lr){3-6}\cmidrule(lr){7-10}
 &  & ViT-Ti & ViT-S & ViT-B & ViT-L & ViT-Ti & ViT-S & ViT-B & ViT-L \\
\midrule
ViT-Ti & 56.30$\pm$0.24 & 56.11$\pm$0.54 & 57.10$\pm$0.44 & 57.82$\pm$0.37 & 58.61$\pm$0.79 & 56.07$\pm$0.62 & 57.61$\pm$0.61 & 57.95$\pm$0.74 & 58.92$\pm$0.74 \\
ViT-S  & 65.55$\pm$0.17 & 65.20$\pm$0.57 & 65.80$\pm$0.38 & 66.59$\pm$0.44 & 67.00$\pm$0.79 & 65.48$\pm$0.83 & 65.69$\pm$0.89 & 66.79$\pm$0.91 & 67.19$\pm$0.57 \\
ViT-B  & 74.14$\pm$0.25 & 74.11$\pm$0.24 & 74.12$\pm$0.36 & 74.89$\pm$0.39 & 75.16$\pm$0.31 & 74.07$\pm$0.56 & 74.17$\pm$0.79 & 74.56$\pm$0.42 & 75.37$\pm$0.56 \\
ViT-L  & 77.69$\pm$0.19 & 77.91$\pm$0.68 & 77.66$\pm$0.77 & 77.82$\pm$0.53 & 77.91$\pm$0.74 & 77.67$\pm$0.85 & 77.73$\pm$0.82 & 77.73$\pm$0.74 & 77.68$\pm$0.76 \\
\bottomrule
\end{tabular}%
}

\vspace{0.75em}

\resizebox{\textwidth}{!}{%
\begin{tabular}{lccccccccccc}
\toprule
\multicolumn{12}{c}{\textbf{ImageNet Top-1 Accuracy} ($\uparrow$): \textbf{ConvNeXtV2}} \\
\midrule
Target & Baseline & \multicolumn{5}{c}{CKA guide $\rightarrow$ target} & \multicolumn{5}{c}{DiffKNN guide $\rightarrow$ target} \\
\cmidrule(lr){3-7}\cmidrule(lr){8-12}
 &  & CNXV2-Atto & CNXV2-Nano & CNXV2-Tiny & CNXV2-Base & CNXV2-Large & CNXV2-Atto & CNXV2-Nano & CNXV2-Tiny & CNXV2-Base & CNXV2-Large \\
\midrule
CNXV2-Atto  & 71.25$\pm$0.11 & 71.59$\pm$0.56 & 71.80$\pm$0.49 & 72.89$\pm$0.51 & 73.93$\pm$0.47 & 74.24$\pm$0.51 & 71.61$\pm$0.67 & 71.80$\pm$0.92 & 72.95$\pm$0.83 & 74.00$\pm$0.42 & 74.49$\pm$0.82 \\
CNXV2-Nano  & 73.14$\pm$0.16 & 73.10$\pm$0.57 & 73.11$\pm$0.34 & 74.46$\pm$0.39 & 74.98$\pm$0.56 & 76.01$\pm$0.61 & 73.19$\pm$0.65 & 73.09$\pm$0.36 & 74.61$\pm$0.41 & 75.12$\pm$0.38 & 76.20$\pm$0.57 \\
CNXV2-Tiny  & 79.41$\pm$0.19 & 79.42$\pm$0.77 & 79.41$\pm$0.39 & 79.73$\pm$0.42 & 80.99$\pm$0.36 & 81.22$\pm$0.47 & 79.34$\pm$0.82 & 79.32$\pm$0.76 & 79.73$\pm$0.77 & 80.96$\pm$0.82 & 81.26$\pm$0.77 \\
CNXV2-Base  & 81.63$\pm$0.15 & 81.01$\pm$0.57 & 81.04$\pm$0.58 & 81.60$\pm$0.48 & 81.82$\pm$0.47 & 82.59$\pm$0.76 & 81.02$\pm$0.74 & 81.01$\pm$0.51 & 81.21$\pm$0.46 & 81.68$\pm$0.47 & 82.71$\pm$0.87 \\
CNXV2-Large & 83.76$\pm$0.18 & 83.71$\pm$0.26 & 83.72$\pm$0.54 & 83.76$\pm$0.55 & 83.74$\pm$0.79 & 83.80$\pm$0.68 & 83.64$\pm$0.33 & 83.68$\pm$0.45 & 83.77$\pm$0.37 & 83.71$\pm$0.75 & 83.79$\pm$0.91 \\
\bottomrule
\end{tabular}%
}

\vspace{0.75em}

\resizebox{\textwidth}{!}{%
\begin{tabular}{lccccccccccc}
\toprule
\multicolumn{12}{c}{\textbf{Perplexity} ($\downarrow$): \textbf{GPT-2}} \\
\midrule
Target & Baseline & \multicolumn{5}{c}{CKA guide $\rightarrow$ target} & \multicolumn{5}{c}{DiffKNN guide $\rightarrow$ target} \\
\cmidrule(lr){3-7}\cmidrule(lr){8-12}
 &  & GPT-2-Nano & GPT-2-Tiny & GPT-2-Base & GPT-2-Medium & GPT-2-Large & GPT-2-Nano & GPT-2-Tiny & GPT-2-Base & GPT-2-Medium & GPT-2-Large \\
\midrule
GPT-2-Nano   & 69.11$\pm$0.96 & 69.23$\pm$1.17 & 68.08$\pm$1.15 & 66.12$\pm$1.05 & 65.87$\pm$1.07 & 65.89$\pm$1.06 & 69.17$\pm$0.89 & 68.06$\pm$1.01 & 66.03$\pm$0.92 & 65.54$\pm$0.96 & 65.49$\pm$0.99 \\
GPT-2-Tiny   & 51.16$\pm$0.91 & 51.32$\pm$1.06 & 51.05$\pm$1.07 & 48.69$\pm$1.03 & 48.65$\pm$1.07 & 48.42$\pm$1.05 & 51.14$\pm$1.10 & 51.19$\pm$0.93 & 48.55$\pm$0.96 & 48.51$\pm$1.00 & 48.53$\pm$1.01 \\
GPT-2-Base   & 28.01$\pm$0.93 & 29.02$\pm$1.02 & 28.16$\pm$1.01 & 28.17$\pm$1.07 & 26.91$\pm$1.08 & 26.87$\pm$1.02 & 28.56$\pm$0.95 & 28.04$\pm$1.04 & 27.99$\pm$1.05 & 26.74$\pm$1.06 & 26.70 $\pm$0.97\\
GPT-2-Medium & 22.55$\pm$0.98 & 23.98$\pm$1.01 & 22.51$\pm$1.08 & 22.52$\pm$1.07 & 22.41$\pm$1.00 & 21.87$\pm$1.00 & 24.14$\pm$0.98 & 22.64$\pm$0.97 & 22.57$\pm$1.04 & 22.51$\pm$1.05 & 21.78$\pm$1.10 \\
GPT-2-Large  & 16.44$\pm$0.90 & 16.91$\pm$0.96 & 16.93$\pm$1.10 & 16.76$\pm$1.09 & 16.65$\pm$1.02 & 16.42$\pm$1.01 & 17.10$\pm$0.92 & 16.42$\pm$0.93 & 16.41$\pm$1.01 & 16.59$\pm$1.02 & 16.40$\pm$0.91 \\
\bottomrule
\end{tabular}%
}

\caption{\small \textbf{NoT transfers across model families and scales.} Each row is a target network, and each guide column reports the resulting NoT performance when transferring from that guide into the target. The diagonal entries correspond to self-transfer. Baseline denotes the target-only model performance. Vision results report ImageNet top-1 accuracy, where higher is better; language results report perplexity, where lower is better.}
\label{tab:scale_transfer_compact}
\end{table*}

\section{Theoretical Results}
\label{ap:theory}

This section explains why an initialization-only alignment procedure can affect
early supervised training. After alignment, the guide is
removed, so the only object that can influence later learning is the geometry
encoded in the target initialization. On a supervised batch $(X,Y)$, this geometry
is captured by the target feature kernel
\[
K_0^S(X)=H_0^S(X)H_0^S(X)^\top .
\]
We compare it to the corresponding guide kernel
\[
K_0^D(X)=H_0^D(X)H_0^D(X)^\top
\]
through
\[
\varepsilon_0(X)=\|K_0^S(X)-K_0^D(X)\|_{\mathrm{op}} .
\]
This derivation necessarily is conditional on this quantity being small because
alignment is optimized on the unlabeled set $U$, while supervised learning happens
on batches $X$.

\paragraph{Connecting alignment to kernel similarity.}
CKA alignment controls the normalized centered kernel shape. In particular, if two
nonzero kernels $A$ and $B$ satisfy
\[
\left\langle \frac{A}{\|A\|_F}, \frac{B}{\|B\|_F}\right\rangle_F = 1-\delta,
\]
then
\[
\left\|
\frac{A}{\|A\|_F}
-
\frac{B}{\|B\|_F}
\right\|_F
=
\sqrt{2\delta}.
\]
Thus, high CKA means that the two kernels have similar shape, up to scale. If their
norms are also close, then the raw kernels are close. This clarifies what CKA does
and does not guarantee: it aligns normalized geometry, while supervised dynamics
depend on the raw kernel or NTK.

\paragraph{First supervised step.}
Consider logits
\[
F=H_\theta(X)V
\]
with a zero-initialized head $V_0=0$. Let
\[
G_0=\nabla_F\mathcal L(F_0,Y).
\]
Since $F_0=0$, the first backbone update is zero. The gradient through the backbone
is multiplied by $V_0$. The readout update is
\[
V_1=-\eta H_0^\top G_0,
\]
and therefore
\[
F_1=H_0V_1=-\eta H_0H_0^\top G_0=-\eta K_0G_0 .
\]
Applying this identity to the aligned target $S$ and the guide $D$ gives
\[
\|F_1^S-F_1^D\|_F
\le
\eta
\|K_0^S-K_0^D\|_{\mathrm{op}}
\|G_0\|_F
=
\eta \varepsilon_0(X)\|G_0\|_F .
\]
Thus, if alignment makes the target and guide kernels close, then their first
supervised logit updates are close.

\paragraph{Initial NTK interpretation.}
At zero head, the empirical NTK of
\[
f_{\theta,V}(x)=V^\top h_\theta(x)
\]
is exactly
\[
\Theta_0=K_0\otimes I_C .
\]
The backbone contribution vanishes because $V_0=0$, and the head contribution is
the feature kernel replicated across classes. Therefore, matching the initial
feature kernels is the same as matching the initial NTKs in this setting.

\paragraph{Short-time dynamics.}
The same conclusion holds for the beginning of gradient flow. With $V_0=0$,
\[
V_t=-tH_0^\top G_0+o(t),
\qquad
\theta_t=\theta_0+o(t),
\qquad
F_t=-tK_0G_0+o(t)
\]
as $t\downarrow 0$. Therefore the initial kernel controls not only the first
discrete step, but also the leading-order continuous-time trajectory.

\paragraph{Robustness and finite time.}
The zero-head assumption is mainly for clarity. If the head is small,
$\|V_0\|_F\le \alpha$, then the initial backbone gradient is small and the NTK is
only a small perturbation of $K_0\otimes I_C$:
\[
\|\nabla_\theta \mathcal L(F_0,Y)\|
\le
B_X\alpha\|G_0\|_F,
\qquad
\|\Theta_0-K_0\otimes I_C\|_{\mathrm{op}}
\le
B_X^2\alpha^2 .
\]
For longer times, one can compare the aligned target trajectory to a hypothetical
guide trajectory. If both NTKs remain close to their initial values and the loss
gradient is Lipschitz in the logits, then the logit distance satisfies a Gronwall
bound:
\[
\delta_t
\le
e^{\kappa Lt}\delta_0
+
G\int_0^t e^{\kappa L(t-s)}
\big(\epsilon_0+\rho_s^S+\rho_s^D\big)\,ds .
\]
This bound says that when the initial kernels are close and the NTKs do not drift too quickly, the early supervised trajectories remain close.

\paragraph{Takeaway.}
The optimization prior does not transfer labels, logits, or trained knowledge.
It changes the target initialization so that the target begins supervised learning
with a feature kernel closer to that of the guide. This kernel determines the first
supervised updates, and under mild stability assumptions it also explains the
early training trajectory.
\section{Recursive Transfer} 
\label{ap:recursive}
\begin{figure}[h]
    \centering
    \includegraphics[width=\textwidth]{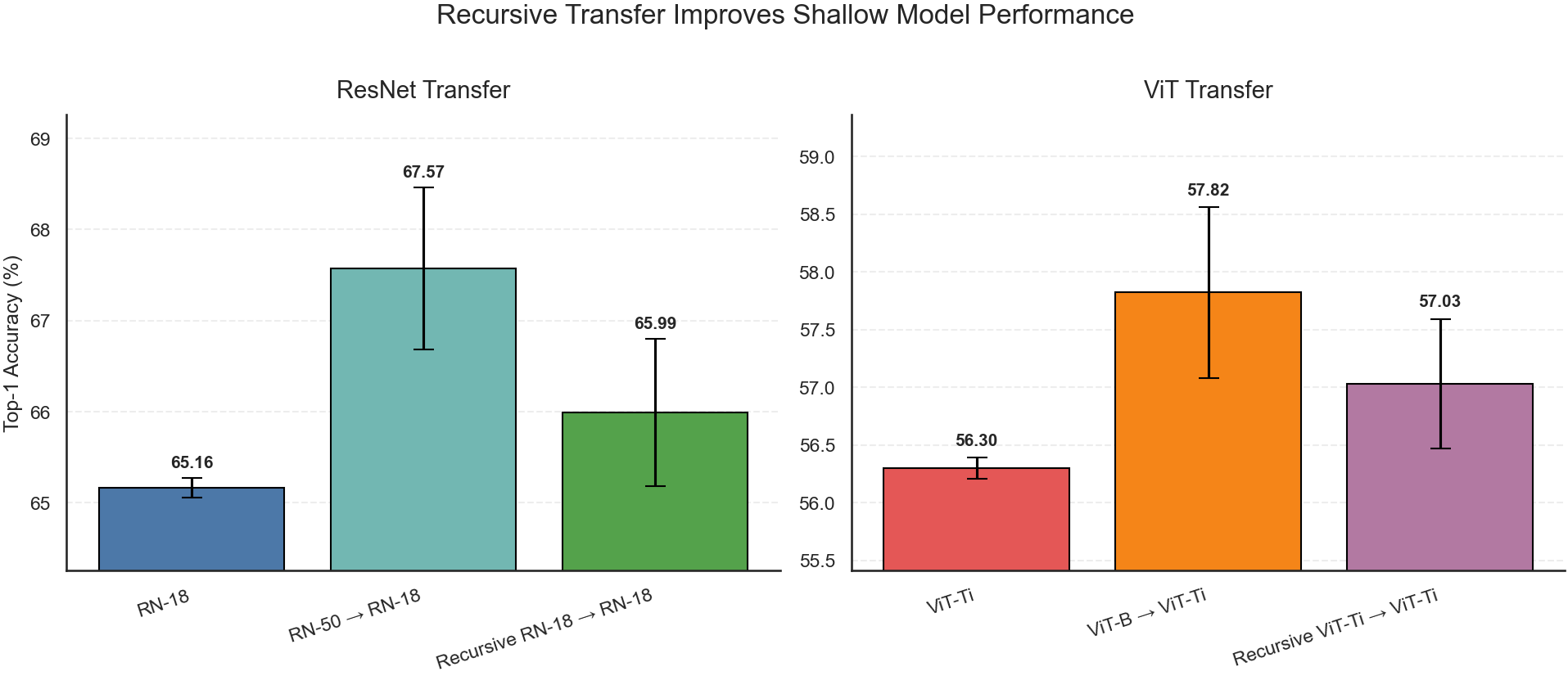}
    \caption{\small \textbf{Recursive guide networks can improve shallow network performance.} We construct a randomly initialized recursive guide by repeating a single block from the target architecture, matching the depth of a larger guide such as ResNet-50 or ViT-B. Recursive guides improve ResNet-18 and ViT-Ti over standard initialization, although they are weaker than the original deep guides. This suggests that additional guide computation helps, but width, spatial structure, and other architectural factors also matter.}
    \label{fig:recursive_transfer}
\end{figure}

To test whether additional guide computation is sufficient to improve the target, we introduce guide networks that apply recursive computation.
This follows previous work \citep{yang2023looped, xu2024expressive, fan2024looped} showing that recursive or looped computations can improve model performance.
We take a single block of the target network and repeat it several times, applying the same block recursively to its own activations.
For ResNets, we repeat a single ResNet-18 block from the first stage to construct a recursive guide with the same depth as ResNet-50.
For ViTs, we repeat a single ViT-Ti block to construct a recursive guide with the same depth as ViT-B.
We show results in Figure~\ref{fig:recursive_transfer}.
We find that randomly initialized recursive guides improve shallow target performance, although not as strongly as the original deep guide networks.
This indicates that additional computation in the guide can help induce better target initializations.
At the same time, the gap between recursive guides and full deep guides suggests that depth alone is not the whole story.
Other architectural factors, including width and spatial structure, also contribute to the geometry transferred by the guide.
\label{ap:alignment_ablation}
\begin{figure}
    \centering
    \includegraphics[width=0.8\textwidth]{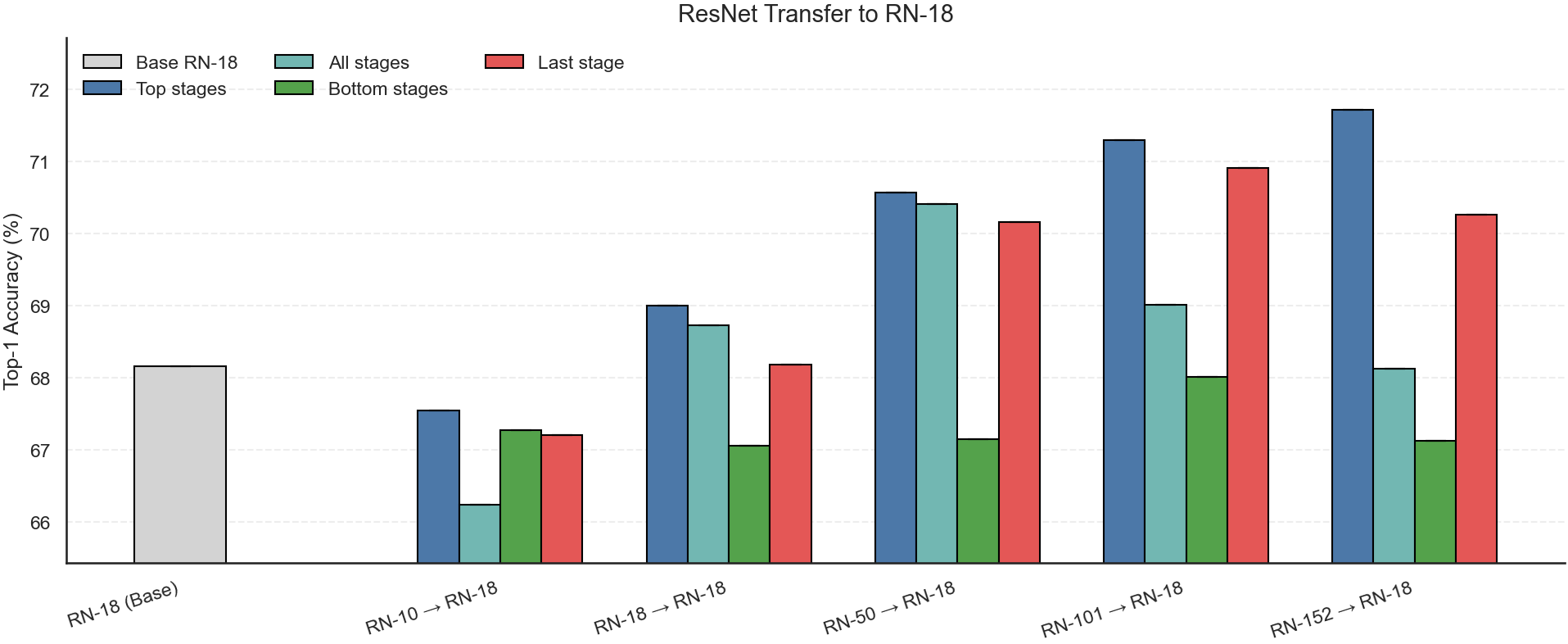}
    \caption{\textbf{Selecting the top half of NoT stages yields the best results.} Rather than aligning all stages across our target and guide network, in this work, we only align the top half of stages in the target network to the randomly initialized guide network. Above, we compare this choice with aligning all the stages, the bottom half of stages, or just the last stage. Only the last stage achieves comparable results but falls short when transferring ResNet-152 to ResNet-18. This establishes that the top half of stages have enough rich geometry to differentiate across networks. A similar trend exists for instilling detail to coarse representations.}
    \label{fig:ablation}
\end{figure}

\section{Ablation Experiments}

As discussed in Section~\ref{sec:experiments}, we use a budgeted version of NoT in comparison to what is reported in prior work \citep{subramaniam2025network}. 
Early experiments established that using all stages in NoT led to noisy results in most settings with instilling depth and instilling detail. 
Upon further inspection, we found that most networks had very similar representations at early stages. 
This was not found in previous works as representations were fairly different across the target and guide network settings. 

In this work, settings like transferring ResNet-101 to ResNet-18 would fail when using all stages for alignment. 
To test this thoroughly, we ran a systematic ablation where we compared using all stages of NoT with using half the stages and using only the last stage for alignment. 
We report results in Figure~\ref{fig:ablation} where we show our results with ResNet-18 as the target network and vary the ResNet depth as the guide network.
We find that results are noisy and inconsistent when using all stages of NoT, so early layers of ResNet-18 are aligned to early layers of the randomly initialized guide network.
For example, we can faithfully improve when aligning ResNet-18 to ResNet-50 but do not succeed when aligning to ResNet-101. This is confirmed when using only the bottom half of stages for alignment, which achieves much worse results when aligning ResNet-18 to ResNet-101. When aligning based on only the last stage, we achieve slightly worse results than using the top half of the stages. We speculate that in cases with less compute, using the last stage would be sufficient in certain settings. We find a similar trend when analyzing our setting aligning ViT-B/32 to ViT-B/16, indicating that the early layers of two networks are often very similar. 


\end{document}